\documentclass{Interspeech}
\interspeechcameraready

\title{Articulatory strategy in vowel production as a basis for speaker discrimination}

\author[affiliation={1}]{Justin J. H.}{Lo}
\author[affiliation={2}]{Patrycja}{Strycharczuk}
\author[affiliation={1}]{Sam}{Kirkham}

\affiliation{Linguistics and English Language}{Lancaster University}{United Kingdom}
\affiliation{Linguistics and English Language}{University of Manchester}{United Kingdom}
\email{\{j.h.lo|s.kirkham\}@lancaster.ac.uk, patrycja.strycharczuk@manchester.ac.uk}

\newcommand{\cllr}{\ensuremath{C\textsubscript{llr}}}
\usepackage{comment}

\begin{document}
\maketitle

\begin{abstract}
    The way speakers articulate is well known to be variable across individuals while at the same time subject to anatomical and biomechanical constraints. In this study, we ask whether articulatory strategy in vowel production can be sufficiently speaker-specific to form the basis for speaker discrimination. We conducted Generalised Procrustes Analyses of tongue shape data from 40 English speakers from the North West of England, and assessed the speaker-discriminatory potential of orthogonal tongue shape features within the framework of likelihood ratios. Tongue size emerged as the individual dimension with the strongest discriminatory power, while tongue shape variation in the more anterior part of the tongue generally outperformed tongue shape variation in the posterior part. When considered in combination, shape-only information may offer comparable levels of speaker specificity to size-and-shape information, but only when features do not exhibit speaker-level co-variation.
\end{abstract}

\section{Introduction}

Articulatory strategy is characterised by individual variation, such that different speakers may perform different articulatory movements to achieve similar acoustic outputs \cite{johnson1993, mielke2016}. Such variation arises as a result of an interplay between speaker anatomy and speaker choice. Differences in the exact shape of every individual's vocal tract might produce acoustic differences in and of themselves, and may also influence the speaker to produce a specific articulatory behaviour in order to achieve a specific acoustic output \cite{nolan1983}. For example, palate morphology can systematically affect the articulatory strategy for the production of both vocalic and consonantal phonemes \cite{brunner2009, rudy2013, weirich2013}. Anatomical factors, however, are not deterministic, as evidenced, for instance, by articulatory differences between monozygotic twins \cite{weirich2012}. Speakers may choose one strategy over another as a result of their individual language experience, which is an additional source of articulatory variation. Furthermore, reliance on a particular strategy is potentially systematic within speaker, due to the influence of a wider articulatory setting \cite{laver1979}, and manifesting as reliance on particular motor routines in speech production that is characteristic of an individual.

At the same time, variation in articulatory strategy is constrained by the broad similarities in the shape of the human vocal tract, by articulation--acoustics relationships, and by biomechanical constraints on articulatory movement, such that key aspects of articulation can be captured using a limited number of parameters \cite{johnson1993, harshman1977, stone1997, story1998}. Furthermore, articulatory setting may attain higher-level regularity through social processes, such that members of the same groups converge on a similar set of articulatory behaviours \cite{gold-2022, knowles-1978}.

The tension between multiple sources of individuality in articulatory strategy and the regularising factors raises the question of whether articulatory strategy is in fact speaker-specific. Can individual speakers be discriminated based on the characteristics of the vocal tract postures they employ to produce contrasts in speech? Establishing the individuality of such differences could contribute to the explanations of the phonetic sources of individuality in human speech. 

To this end, we investigate whether tongue shape features in vowel production can be the basis of successful speaker discrimination. Vowels have long been of considerable interest to forensic phoneticians for their potential as speaker discriminants \cite{fairclough-2023}, with long-term distributions of formants (LTFDs) in particular offering the capacity to reflect both individual physiology and articulatory habit \cite{nolan-2005}. We take a similar approach to examine articulation of the whole vowel space, with the aim of inquiring into variability and individuality in broad articulatory tendencies. In a larger perspective, acoustic models of vowels going back to \cite{stevens1955} and \cite{fant1971} provide a framework for linking individuality in production with individuality in acoustics. Due to the nature of our data, which come from ultrasound tongue imaging, we focus on the tongue. Ultrasound does not capture direct information about jaw and lip displacement and, as such, this approach offers only a partial view of articulation, yet it provides detailed information on tongue movement, which is a crucial component in vowel production.

Our study aims to address the following research questions:

\begin{enumerate}
    \item Do tongue shape features used in vowel production contain speaker-specific information?
    \item Which tongue shape features are the most informative for speaker discrimination and can therefore be considered individual?
\end{enumerate}

We apply the framework of likelihood ratios (LRs) to assess speaker specificity. A modern standard in forensic voice comparison, LRs are used to evaluate the strength of voice evidence under competing same- and different-speaker hypotheses. Similarities of features between voice samples are thus considered in the light of their typicality within the relevant population. While articulatory information is not directly accessible in forensic casework, LR-based testing provides a common framework for the speaker-discriminatory potential of articulatory features to be compared with their acoustic counterparts.

\section{Tongue size \& shape features}

\subsection{Ultrasound tongue imaging data}

Our study includes data from 40 speakers of English from the North West of England (Greater Manchester and Lancashire). Midsagittal ultrasound tongue imaging data were collected from these speakers while they produced a full set of stressed vowel phonemes in a fixed segmental environment (/bV/ and /bVd/, e.g., \emph{bee, bead, bar, bard}). Note that Northern English is typically non-rhotic and has no \textipa{/U/}--\textipa{/2/} split. The data were collected using the EchoB system (10 speakers) and the Telemed Micro ultrasound speech system (30 speakers) from Articulate Instruments Ltd. The sampling rate of the ultrasound system ranged between 59.5 and 101 frames per second (median $=$ 81.3 fps). Headset stabilisation was used to minimise probe movement. Typically 4--6 repetitions of the test stimuli were obtained from each speaker. The speakers read the items from a screen, embedded within a carrier phrase, and were instructed to speak as naturally as possible. The study received ethical approval from The University of Manchester and Lancaster University.

The ultrasound data were processed using DeepLabCut \cite{mathis2018}, as implemented within Articulate Assistant Advanced, version 2.20 \cite{wrench2022}. DeepLabCut uses the pre-trained unidirectional MobileNet1.0 model, version 1.1.0, to automatically label 11 points corresponding to key anatomical landmarks on the midsagittal tongue contour, which were exported as Cartesian coordinates (in \unit{\milli\metre}) and rotated on the occlusal plane. For the purpose of the current paper, we focus on the data corresponding to the acoustic midpoint of the vowel.

To remove erroneous landmark identification by DeepLabCut, we calculated the median coordinates of each landmark for every speaker, then excluded trials with any landmarks over 3.5 median absolute deviations away by Euclidean distance from the speaker's median landmarks. 4.5\% of trials were removed as a result, leaving 10,407 trials remaining for analysis.

\subsection{Shape analysis}\label{sec:gpa}

While landmark coordinates hold information about the configuration of the tongue, they are susceptible to translational and rotational effects introduced by variation in probe placement and standardisation with reference to each speaker's own occlusal plane. Such methodological influences have to be factored out from the data, in order to isolate lingual anatomy and articulatory behaviour as sources of speaker variability. Since tongue shape measures tend to be more robust against probe displacement than displacement measures \cite{menard2012}, we focus on tongue shape features derived using existing general methods for statistical shape analysis \cite{shapes-gpa}.

We used the \emph{shapes} package \cite{shapes} in R to apply Generalised Procrustes Analysis (GPA) to the set of tongue contours, which performed normalisation by translation and rotation to minimise the Procrustes distance between the contours. We applied two separate procedures: (1) A partial GPA which did not involve scaling the data and so preserved information about both tongue size and tongue shape; and (2) a full GPA with scaling that preserved only tongue shape. We then performed a principal components (PC) analysis in the tangent space \cite{kent1992} to identify the main orthogonal dimensions of variation. We focus on the first three PCs, which together explained 90.6\% of the variance in the size-and-shape analysis and 88.3\% in the shape analysis.

\begin{figure}[ht]
  \centering
  \includegraphics[width=\linewidth]{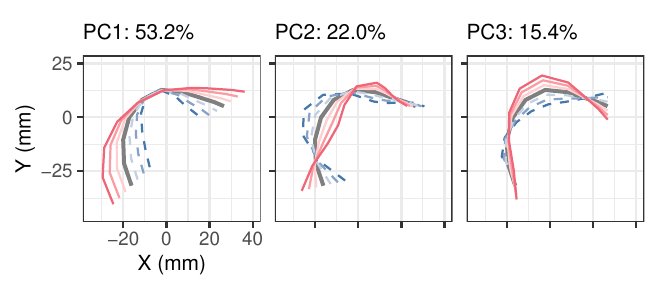}
  \caption{Effects of PC1--3 (left to right) in the size-and-shape PCA. Each panel shows the mean shape (heavy grey) and variation up to 3 SDs higher (solid red) and lower (dashed blue).}
  \label{fig:unscaled-pca}
\end{figure}

Figure \ref{fig:unscaled-pca} summarises the effects of the first three PCs in the \emph{size-and-shape} analysis. PC1 is primarily related to tongue size, with higher PC1 scores indicating larger tongues. This interpretation is supported by the strong correlation between PC1 scores and the centroid size of the tongue splines (Pearson's $r=0.972$, $p < 0.0001$). PC2 captures variation in tongue curvature centred at the tongue root, where greater convexity is represented by low PC2 scores. High PC2 scores correspond to a degree of concavity at the same location, which is also associated with bunching of the front part of the tongue dorsum, such that the overall location of tongue curvature is more anterior. PC3 is similarly associated with variation in the degree of curvature, but in the tongue dorsum region. High PC3 scores correspond to dorsum arching, whereas low PC3 scores produce an overall flatter tongue shape.

\begin{figure}[ht]
  \centering
  \includegraphics[width=\linewidth]{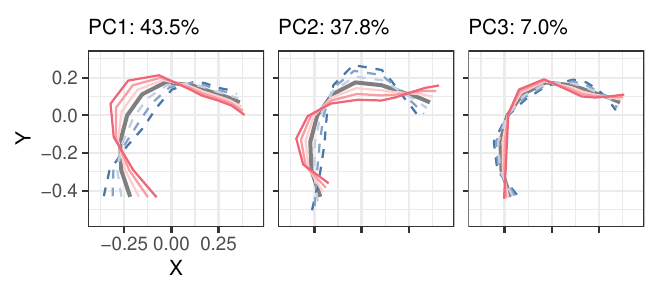}
  \caption{Effects of PC1--3 (left to right) in the shape PCA. Each panel shows the mean shape (heavy grey) and variation up to 3 SDs higher (solid red) and lower (dashed blue).}
  \label{fig:scaled-pca}
\end{figure}

The effects of the first three PCs in the \emph{shape} analysis are summarised in Figure \ref{fig:scaled-pca}. All three PCs can be characterised as representing variation in the degree of tongue curvature, centred at progressively more anterior parts of the tongue. PC1 targets the posterior part of the tongue dorsum, where high PC1 scores correspond to its arching and low PC1 scores result in a slight concavity at the same location. PC2 addresses the anterior part of the tongue dorsum: Low PC2 scores correspond to this part of the tongue being more convex and raised, whereas high PC2 scores correspond to a more concave and flattened anterior part of the tongue dorsum. It should be pointed out that, although shape PC1 and PC2 primarily affect the dorsum as in the case of size-and-shape PC3, they are not focused on the same location. In particular, shape PC1 and PC2 are centred on each side of the focal landmark of size-and-shape PC3. For both shape PC1 and PC2, concavity at the respective part of the tongue dorsum has the effect of shifting the location of main tongue curvature elsewhere, forward towards the tongue mid for PC1, and backward towards the tongue root for PC2. Shape PC3 targets a more anterior part of the tongue and captures variation from a convex (low PC3 scores) to concave (high PC3 scores) tongue mid. The tongue shape associated with high PC3 scores -- raised tongue dorsum and concave tongue mid -- is reminiscent of a velarised sound such as [\textltilde] (cf. speakers S1 and S3 in \cite[Fig. 4]{strycharczuk_kiwi} for examples).

In both \emph{size-and-shape} and \emph{shape} analyses, individual PCs capture tongue shape variation in closely adjacent parts of the tongue. We can thus expect, in particular, size-and-shape PC2 and PC3, or shape PC1 and PC2, to interact in manoeuvring the location of main tongue curvature and producing vowel contrasts. High front vowels like /i/, for example, involve a depression in the posterior part of the tongue and arching of the anterior part of the dorsum, which can be achieved through a combination of high size-and-shape PC2 and PC3 scores or low shape PC1 and PC2 scores.

\subsection{Predictions for speaker discrimination}

Given that PC1 in the size-and-shape analysis is mainly about tongue size, we expect it to encode speaker-idiosyncratic information and in principle perform well as a speaker discriminant. Meanwhile, as size-and-shape PC2 and PC3 are primarily governed by vowel identity, they are predicted to exhibit high within-speaker variability and low individuality. By the same token, shape PC1 and PC2 are not expected to possess strong speaker specificity. Shape PC3, on the other hand, is potentially more variable, as it is not one of the two primary dimensions for capturing vowel contrasts. Furthermore, as previously mentioned, high PC3 scores represent a tongue shape in which the anterior part of the tongue is concave and the dorsum is raised, seen, for instance, in a velarised [\textltilde]. Example data in \cite{strycharczuk_kiwi} suggest that, in liquid production, this feature can be somewhat idiosyncratic. Assuming that similar variation exists for back vowels, we can expect a degree of individuality in this respect.

\section{Likelihood ratio-based testing}

To assess speaker individuality of tongue size and shape in vowel articulation, we tested how well the PCs derived in Section \ref{sec:gpa} performed as speaker discriminants using LRs. We tested two sets of systems using PCs from the size-and-shape analysis and the shape analysis as input features respectively. Within each set, we tested the performance of each individual PC1--3 as well as all their possible combinations.

We performed system testing with the \emph{fvclrr} package \cite{fvclrr} in R using a cross validation procedure. For each combination of input features, we compared the first half of each speaker's data with the second half from each of the 40 speakers using the multivariate kernel density formula \cite{aitken-lucy-2004}. Typicality was assessed with reference to all speakers, excluding the two speakers being compared in different-speaker comparisons, or the target speaker plus another random speaker in same-speaker comparisons. Splitting the data in half allowed us to conduct same-speaker comparisons using distinct data from a single recording session, although we acknowledge that this likely underestimates the extent of within-speaker variability. As speakers recorded repeated blocks of stimuli, the subsets were balanced in terms of vowel distribution. In total, these pairwise comparisons produced 1,600 scores per system, which were then calibrated by logistic regression \cite{morrison-2013} to produce log\textsubscript{10}LRs.

System performance was evaluated using two standard metrics: Equal error rate (EER), aimed at quantifying the proportion of contrary-to-fact LRs, and the log LR cost ({\cllr}) \cite{brummer-dupreez-2006}, which also reflects the magnitude of any errors. Systems with $\cllr < 1$ are considered well-calibrated and contain useful speaker-specific information, and stronger performance is indicated by values of EER and {\cllr} closer to $0$.

\section{Results}\label{sec:results}

\begin{figure}[ht]
  \centering
  \includegraphics[width=0.9\linewidth]{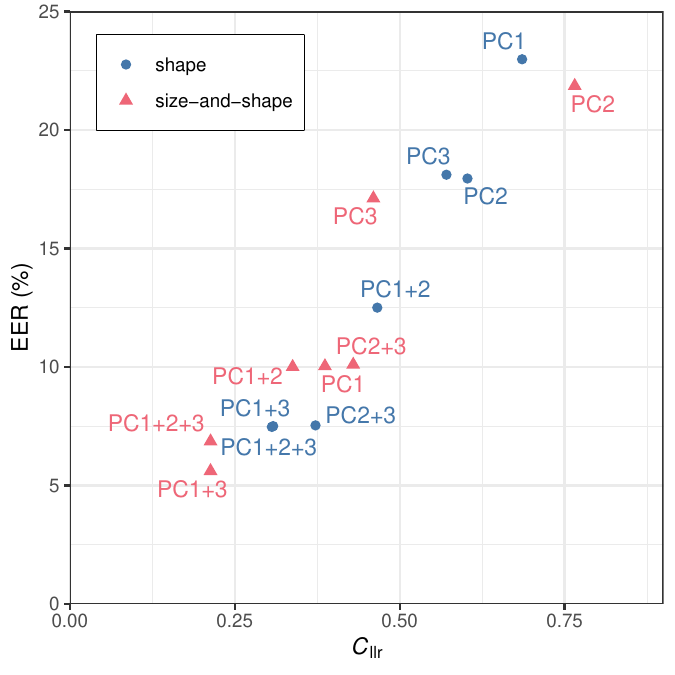}
  \caption{Performance of systems with size-and-shape (blue circles) and shape PCs (red triangles) as input.}
  \label{fig:metrics}
\end{figure}

System performance, as evaluated by EER and {\cllr}, is reported in Table \ref{tab:metrics} and illustrated in Figure \ref{fig:metrics}. All systems tested produced {\cllr} below $1$, meaning that each combination of PCs was able to capture some speaker-specific information.

\begin{table}[ht]
  \caption{EER and {\cllr} from all systems tested}
  \label{tab:metrics}
  \centering
  \begin{tabular}{l S[table-format=2.1] S[table-format=1.3] S[table-format=2.1] S[table-format=1.3]}
    \toprule
    & \multicolumn{2}{c}{\textbf{Size-and-shape}} & \multicolumn{2}{c}{\textbf{Shape}} \\
    & {EER (\%)} & \cllr & {EER (\%)} & \cllr \\
    \midrule
    PC1     & 10.0 & 0.387 & 23.0 & 0.686 \\
    PC2     & 21.9 & 0.765 & 17.9 & 0.603 \\
    PC3     & 17.1 & 0.460 & 18.1 & 0.571 \\
    PC1+2   & 10.0 & 0.338 & 12.5 & 0.466 \\
    PC1+3   & 5.6 & 0.213 & 7.5 & 0.308 \\
    PC2+3   & 10.1 & 0.430 & 7.5 & 0.372 \\
    PC1+2+3 & 6.9 & 0.213 & 7.5 & 0.306 \\
    \bottomrule
  \end{tabular}
\end{table}

Among systems using PCs from the \emph{shape} analysis as input, those using only a single PC all performed relatively poorly, with EERs of around 18\% or above. PC1 performed the worst, with the highest EER and {\cllr}, while PC3 slightly outperformed PC2 with a lower {\cllr}. Combinations of two PCs resulted in improved system performance over their individual components, with the combination of PC1 and PC3 producing the lowest EER and {\cllr}, followed closely by the combination of PC2 and PC3. However, the system with all three PCs as input did not result in further increase in performance, but instead yielded the same EER as the PC1+3 system and a negligibly lower {\cllr}.

As for the set of systems using PCs from the \emph{size-and-shape} analysis, PC1 was the individual PC that produced the system with the lowest EER and {\cllr}. Both PC1 and PC3 in this set outperformed all individual shape PCs, whereas performance for PC2 was at a similar level to shape PC1. While combining two PCs in this set similarly improved performance upon the component PCs, the combination of PC2 and PC3 performed worse than combinations involving PC1 and even PC1 alone. The combination of PC1 and PC3 produced the strongest performance out of all systems tested, with the system using all three PCs as input closely behind, reporting the same {\cllr} and a slightly higher EER than the one with PC2 left out.

\section{Discussion}

We set out to investigate whether tongue shape features in vowel articulation are speaker-specific. The results in Section \ref{sec:results} show that all the features we tested carry some speaker-specific information, although there is substantial variation in their utility for speaker discrimination.

Tongue size, as reflected in size-and-shape PC1, trumped all other tongue shape features, reaffirming the invariant nature of tongue volume. Its primarily biological basis, however, is not directly informative of articulatory strategy, although \cite{johnson2023} observes a correlation between vocal tract length and preference for some articulatory strategies. While the other features were individually weak speaker discriminants, variation in the concavity of the tongue dorsum (size-and-shape PC3) showed stronger speaker specificity, while variation in tongue root shape (size-and-shape PC2) fared poorly. Likewise, the shape of the tongue mid (shape PC3) and the anterior part of the tongue dorsum (shape PC2) provided stronger discriminatory power than that of the posterior part of the tongue dorsum (shape PC1).

It may be the case that the shape of the posterior part of the tongue is relatively weak in speaker specificity because it is primarily determined by the displacement of the tongue root in the front--back dimension of the articulatory space. Horizontal displacement of the tongue root varies systematically with vowel identity, acting as a relatively reliable vowel discriminant, and also shows a strong correlation with F2 \cite{strycharczuk2025}. These characteristics would lead shape PC1 and size-and-shape PC2 to exhibit consistently high within-speaker variability in our data, which would likely have contributed to their ineffectiveness as speaker discriminants. Such a finding also aligns with forensic research on LTFDs, which has shown LTF2 to perform worse than other formants \cite{lo-2021}.

\begin{figure}[ht]
    \centering
    \includegraphics[width=0.8\linewidth]{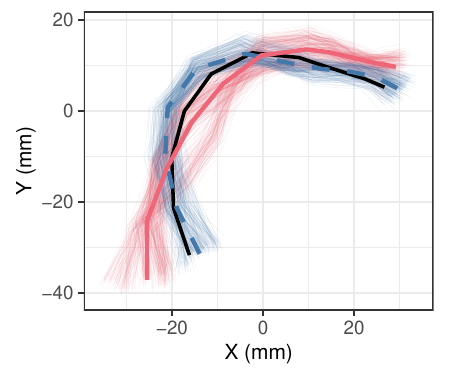}
    \caption{Mean (heavy) and individual (light) tongue shapes of two speakers with the highest (solid red) and lowest (dashed blue) mean PC2 scores in the size-and-shape PCA, together with overall mean shape (solid black).}
    \label{fig:pc2}
\end{figure}

This is not to say that between-speaker variation in overall tendencies of tongue root shape is absent, when in fact speakers with generally advanced or retracted tongue root articulation are both present in our data. Figure \ref{fig:pc2} provides a clear illustration with two speakers with the most extreme mean size-and-shape PC2 scores. It does appear to be the case, however, that there is greater scope for individuality in the shape of the anterior part of the tongue. We believe this can be attributed in part to the interaction between articulatory variability and palate shape, and, consequently, traced back to the idiosyncrasy of palate morphology. Speakers with flatter palates have been shown to be more internally consistent in tongue height, in order to constrain changes to vocal tract area and maintain acoustic constancy \cite{brunner2009}.

Tongue shape features show greater speaker individuality when multiple dimensions are considered in conjunction. While size matters, even tongue shape alone has the potential to convey comparable levels of speaker-specific information in the best case scenario. Our results show that the combination of shape PC1 and PC3, or all three PCs, only slightly underperformed their size-and-shape counterparts. As PCs capture dimensions of variation orthogonal to each other, it is perhaps unsurprising that using multiple PCs would draw upon complementary sources of information to improve speaker discrimination. Yet, in both sets of systems, adding PC2 to the combination of PC1 and PC3 did not further improve performance. Any information about individuality that PC2 had to offer, then, was already captured by PC1 and PC3 together, despite these PCs representing different dimensions in the two sets. One possible explanation is that, while the PCs are orthogonal on the item level, there is speaker-level co-variation between PC2 and the other PCs. Indeed, speaker means of size-and-shape PC2 and PC3 show a weak-to-moderate negative correlation ($r=-0.356$, $p=0.0243$): Speakers with an overall retracted tongue root (low PC2 scores) are also speakers with an overall more arched tongue dorsum (high PC3 scores). Shape PC1 and PC2 are likewise correlated ($r=-0.343$, $p=0.0302$), meaning speakers whose main tongue curvature tends to sit in the posterior part of the tongue dorsum are associated with having a generally flatter shape of the anterior part. There is further speaker-level co-variation in tongue shape variability, as by-speaker standard deviations for shape PC2 and PC3 are positively correlated ($r=0.384$, $p=0.0144$), such that speakers who use a more consistent shape in the tongue mid are also more consistent for the anterior part of the tongue dorsum. These correlations are suggestive of a relationship between articulatory strategy and discriminatory performance on the individual level, which merits further exploration in future work.

\section{Conclusion}

This study marks the first application of LR-based testing to explore speaker-specific information in articulatory features. We found evidence that both anatomical features (tongue size) and articulatory strategy in vowel production (variation in tongue shape) contribute to speaker discrimination. These features, individually or in combination, perform at a level on par with LTFDs \cite{lo-2021, hughes-2017}, which also aim to capture speaker specificity in articulatory setting. While the controlled nature of our materials and differences in data modelling techniques mean that such a comparison needs to be treated with caution, these findings can contribute to disentangling the various sources of individual variation in speech, and ultimately providing a more holistic model of the phonetic basis of speaker discrimination. A crucial step in developing such a model that we plan to undertake will be linking variation in articulatory strategy with specific acoustic features.

\section{Acknowledgements}

This research was supported by an AHRC research grant awarded to PS and SK (AH/S011900/1), a British Academy Mid-Career Fellowship awarded to PS (MFSS24{\textbackslash}240076), and an AHRC Fellowship awarded to SK (AH/Y002822/1).

\bibliographystyle{IEEEtran}
\bibliography{articulatory_individuality}

\end{document}